\documentclass[runningheads]{llncs}
\usepackage{graphicx}
\usepackage{subfigure}
\usepackage{amsmath,amssymb}
\usepackage{color}
\PassOptionsToPackage{hyphens}{url}\usepackage{hyperref}
\usepackage[labelfont=bf]{caption}
\usepackage{amsmath}
\usepackage{booktabs} 
\usepackage{multirow}
\usepackage{float}
\usepackage[dvipsnames]{xcolor}
\usepackage{cite}
\usepackage{soul}

\begin{document}
\title{Multi-stream Fusion for Class Incremental Learning in Pill
Image Classification}
\titlerunning{Multi-stream Fusion for CIL in Pill Image Classification}
\author{Trong-Tung Nguyen\inst{1, 2} \and
Hieu H. Pham\inst{1, 3, *} \and
Phi Le Nguyen\inst{4} \and Thanh Hung Nguyen\inst{4} \and Minh Do\inst{1, 3, 5}}
\authorrunning{Trong-Tung Nguyen et al.}
\institute{VinUni-Illinois Smart Health Center, VinUniversity, Hanoi, Vietnam; \\
\email{\{tung.nt,hieu.ph,minh.do\}@vinuni.edu.vn}\and
John von Neumann Institute, University of Science, VNU-HCM, Vietnam;\and
College of Engineering \& Computer Science, VinUniversity, Hanoi, Vietnam; \and 
School of Information and Communication Technology, Hanoi University of Science
and Technology, Vietnam;\\
\email{\{lenp,hungnt\}@soict.hust.edu.vn}
\and 
University of Illinois at Urbana-Champaign, US;\email{minhdo@illinois.edu}\\
*Corresponding author
}%
\maketitle              % typeset the header of the contribution
\begin{abstract}
Classifying pill categories from real-world images is crucial for various smart healthcare applications. Although existing approaches in image classification might achieve a good performance on fixed pill categories, they fail to handle novel instances of pill categories that are frequently presented to the learning algorithm. 
To this end, a trivial solution is to train the model with novel classes.
However, this may result in a phenomenon known as catastrophic forgetting, in which the system forgets what it learned in previous classes. 
In this paper, we address this challenge by introducing the class incremental learning (CIL) ability to traditional pill image classification systems. 
Specifically, we propose a novel incremental multi-stream intermediate fusion framework enabling incorporation of an additional guidance information stream that best matches the domain of the problem into various state-of-the-art CIL methods. From this framework, we consider color-specific information of pill images as a guidance stream and devise an approach, namely ``\textit{Color Guidance with Multi-stream intermediate fusion}"(CG-IMIF) for solving CIL pill image classification task. We conduct comprehensive experiments on real-world incremental pill image classification dataset, namely VAIPE-PCIL, and find that the CG-IMIF consistently outperforms several state-of-the-art methods by a large margin in different task settings. Our code, data, and trained model are available at \url{https://github.com/vinuni-vishc/CG-IMIF}.

% \keywords{Pill Recognition  \and Class Incremental Learning \and Multi-stream fusion}
\end{abstract}
\section{Introduction}
Pill image recognition task has attracted various studies recently with the aim to design high-quality algorithm for visual-based assistance system on pill images. This can help the healthcare community automatically identify unknown pill categories by taking several real-world pictures with mobile devices. It is noteworthy that real-world scenarios of pill images are often challenging due to the changing background as well as variances of pill instances in terms of shape, color, and texture. There have been several works that are developed to mitigate such challenges, most of them are based on hand-crafted features \cite{6467032, articleB, articleA,articleC}. These works are then utilized by Ling et al.\cite{9157392} and combined with a two-stage training strategy to create a novel framework for the pill recognition model in few-shot learning. Another approach is to explore external knowledge from medical text data (e.g. prescription) to improve the detection performance of visual-based models~\cite{nguyen2022image,nguyen2022novel}. However, existing models are often limited by novel instances of pill categories which frequently arrive at a pill recognition system. This often happens when a novel class of pill instance is introduced by images uploaded from the end-user using mobile devices or from the healthcare community. A report in \cite{fdaCDERsMolecular} shows that there are roughly 40-50 novel drugs being approved each year. 
In such a scenario, the core learning model of the system, which is often deployed in a lightweight device (\textit{e.g}, mobile phones), might need to rewind the training process on the whole training data (in which novel categories participate). This is not an effective strategy for many reasons. Memory allocated for such extensively training data is often limited. Acquiring novel knowledge while maintaining what the model has learned so far requires the system to store a huge amount of samples for both old and new classes, which is infeasible. Another solution for this is to provide an initial training dataset for the model. The model is then fine-tuned on novel categories to update the model's knowledge about new pill instances. However, this fine-tuning scheme suffers from a serious behavior of the learning system which is widely known as catastrophic forgetting \cite{articleCatas, https://doi.org/10.48550/arxiv.1312.6211} (degrading performance on old tasks while accessing data of novel tasks). This system, therefore, is in need of a flexible and effective strategy to handle the novel real-world object categorization of pill image instances. In this way, it would be able to incrementally learn from new classes without exhaustively storing old category samples. This scenario is called class-incremental learning (CIL).

The progress of studies on class incremental learning (CIL) for visual tasks has been developed significantly for many years. The general setting of CIL is that the disjoint sets of different classes arrive at the learning algorithm gradually. Many works such as \cite{DBLP:journals/corr/abs-1903-07864,Hou_2019_CVPR,DBLP:journals/corr/RebuffiKL16,DBLP:journals/corr/abs-1807-09536, DBLP:journals/corr/abs-1905-13260} have proposed several methods which employed available techniques to tackle the mutual challenge: catastrophic forgetting. Knowledge distillation \cite{DBLP:journals/corr/HintonVD15} is the most common technique which is widely adopted to tackle catastrophic forgetting and was first applied to the CIL setting by Li et al.\cite{DBLP:journals/corr/LiH16e}. After that, a derived version \cite{DBLP:journals/corr/RebuffiKL16} with additional usage of representation learning was proposed, in which valuable herding exemplars are replayed frequently to keep track of the old knowledge. The strategy of herding is to pick those neighbors which are nearest to the mean sample of the class. Using this herding strategy, Castro et al.\cite{DBLP:journals/corr/abs-1807-09536} managed to build an end-to-end framework with an additionally balanced fine-tuning strategy. On the other hand, Wu et al.\cite{DBLP:journals/corr/abs-1905-13260} introduced a bias correction approach by adding a bias correction layer. This is conducted at the last layer of each incremental learning task to refine the overall scores for the final prediction. Meanwhile, Hou et al.\cite{Hou_2019_CVPR} identified the imbalance between previous and new data as the main issue leading to catastrophic forgetting. They tackled this imbalanced scenario by incorporating three main components: cosine normalization, less-forget constraint, and inter-class separation.

\begin{figure}[htbp]
    \centering
    \includegraphics[width=0.75\textwidth]{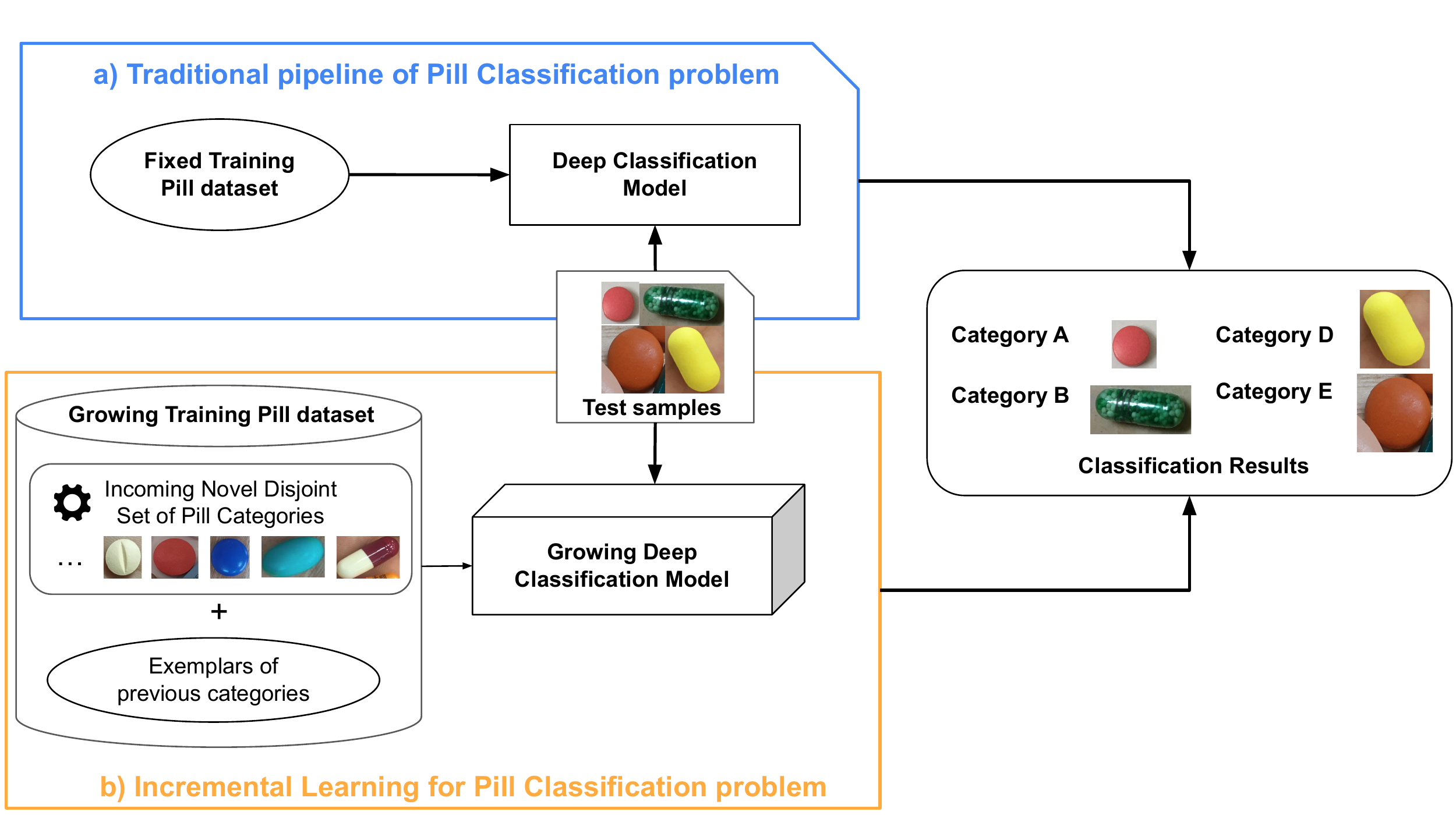}
    \caption{The pipeline for a learning algorithm to acquire knowledge of pill categories could be divided into two options: (a) feeding a fixed pill images database to an off-the-shelf deep learning algorithm; (b) maintaining a few samples of old categories as exemplars, combining with novel categories to form a growing pill image dataset, and finally feeding into a growing deep classification model.}
    \label{fig:overview}
\end{figure}

In this research, we aim to investigate the application of CIL methods in a pill classification system. Fig.\ref{fig:overview} illustrates the effect of such a system with and without class incremental learning capability. To the best of our knowledge, we are the first to explore incremental learning on the pill classification system. Existing single stream incremental learning methods \cite{DBLP:journals/corr/abs-1903-07864,Hou_2019_CVPR,DBLP:journals/corr/RebuffiKL16,DBLP:journals/corr/abs-1807-09536, DBLP:journals/corr/abs-1905-13260}, when being applied to a domain of application for practical usage, can be improved with the help of some domain-specific knowledge. This serves as additional information which might collaborate well with the original RGB image to alleviate catastrophic forgetting. The introduction of a supplementary information stream requires a prudent strategy to incorporate such information. Based on this motivation, we propose a novel integration framework that serves as a plug-in technique for any available class incremental learning algorithms. Our fusion framework enables the incremental learning methods to receive additional information streams as cues. This will then help to flexibly update corresponding feature representations in an optimal way for each learning task through the intermediate stage. To demonstrate the usage of such an integration framework, we consider color information as additional stream and devise an approach, named ``\textit{Color Guidance with Multi-stream intermediate fusion}"(CG-IMIF). Experimental results on a real-world incremental pill image classification dataset called VAIPE-PCIL show that the proposed learning framework consistently surpasses most metric scores of various state-of-the-art methods in different task settings.

\noindent Our contributions can be summarized in the following three aspects: 
\begin{enumerate}
	 \item {We introduce CG-IMIF, a novel incremental learning framework based on multiple streams for the task of pill classification from images. To the best of our knowledge, we are the first to introduce the incremental learning capability to this task and provide a new approach to tackle challenges in learning novel pill classes.}
	 
	 \item {We conduct thorough experiments and in-depth ablation studies to demonstrate the effectiveness of the proposed approach on a real-world incremental pill image classification dataset. Experimental results show that the CG-IMIF consistently outperforms previous state-of-the-art methods by a large margin.}
	% \item {Our codes, trained models and dataset will be made available upon the publication of this paper.}
\end{enumerate}
The rest of this paper is organized as follows. We briefly formulate the problem setting of pill CIL, which we aim to solve in Section 2. Details of our proposed CG-IMIF framework are described in Section 3. Experimental results and further analysis are presented in Section 4 and 5. Finally, we conclude the paper with our discussion on strengths and limitations in Section 6, and 7.
\section{Preliminaries}
\subsection{Problem Definition and Notation}
Generally, the \textbf{Class Incremental Learning (CIL)} problem represented by $\tau$ consists of a sequence of $n$ image classification learning tasks
\begin{equation}
    \tau = [(C^1, P_{train}^1, P_{test}^1), (C^2, P_{train}^2, P_{test}^2), ..., (C^n, P_{train}^n, P_{test}^n)] ,
    \label{equ:probform}
\end{equation}

\noindent where each tuple $(C^t, P_{train}^t, P_{test}^t)$ depicts a task $t$.
$C^t$ is a set of $m^t$ categories, \textit{i.e.}, $C^t=\{c_1^t, c_2^t, ..., c_{m^t}^t\}$, $P_{train}^t$ and $P_{test}^t$ denote the training and testing data, respectively. To represent the total number of classes up to the current task, we define $M^t=\sum_{i=1}^{t}\left | C^i \right |$. The training, and testing data is defined as $P^t = \{(X^t, Y^t)\}$ where $X^t$ and $Y^t$ denote the training images and their corresponding labels, respectively. During the training phase, the learning model at stage $t$ is presented with categories set $C^t$, training samples $P_{train}^t$, and an exemplar set $K_t$. In practice, $K_t$ is a fixed-size set acting as a support set which helps to retain a partial set of images and the corresponding labels from previous training data, \textit{i.e.}, $K_t \subseteq P_{train}^1\cup P_{train}^2 \cup ... \cup P_{train}^{t-1}$. 
Therefore, a revised version of training samples at stage $t$ can be obtained by combining $K_t$ and $P_{train}^t$, $K_t \cup P_{train}^t=V_{train}^t$. It is also assumed that categories of different learning tasks do not overlap (\textit{i.e} $C^i \cap C^j = \varnothing$ where $i \neq j$). At testing time, the performance of learner $t$ is evaluated on all of the previous seen categories $\bigcup_{i=1}^{t}C^t$ with samples from $\bigcup_{i=1}^{t}P_{test}^t$.

\subsection{Conventional CIL Methods}
\label{subsec:conventionalCIL}
Several CIL methods have been proposed which consider various properties of CIL problem to tackle mutual challenge: catastrophic forgetting. Most CIL works are divided into two branches: exemplar-based, and non-exemplar-based approaches. While the latter is much more challenging, the former is more practical since it is reasonable to maintain a few samples of old classes to avoid performance degrading. Within the scope of this research, we aim to exploit the capability of exemplar-based CIL methods by attaching these to our proposed framework. Therefore, we first describe a few core components of exemplar-based CIL approaches as follows.\\
\textbf{Representative Memory} is a set of samples from categories of old tasks and is represented by $K_{t-1}$. It serves as an exemplar set to support model in revisiting knowledge acquired from old tasks. In an exemplar-based approach, the learning model can only access the previous category set $C^{t-1}$ through $K_{t-1}$. The size of the support set is often limited and mainly divided into two memory settings: 1) a constant number of exemplars per class, and 2) and a limited capacity of \textit{S} samples. In the first setting, the size of the support set $K_t$ grows with the number of classes. In addition, the size of $K_t$ in the second memory setting is constant over the time $t$. Samples across categories are manipulated frequently with two main operations: new sample selection and old sample removal. 
For each class, a sorted list of its samples is maintained based on their distances to the class' mean feature vector. Hence, the most representative samples for each class are selected as members in the next support set $K_{t+1}$. Meanwhile, the remaining samples are ignored to reserve slots for novel samples from new classes.\\
\textbf{Growing Deep Neural Networks} in an exemplar-based method is constructed by two main factors: the common feature extractor backbone, and a growing classification layer module. At a specific learning stage $t$, new classification head $CL_t$ is initiated to allocate corresponding parameters $W_t$. Feature vectors, after being extracted by the feature extractor, are fed into $CL_t$ to produce prediction logits for the current category set $C^t$. The size of the logits after being input to $CL_t$ is equal to the size of the category set $C^t$. The prediction vectors are then utilized to compute the traditional cross-entropy loss which represents the training loss on the set of pill images $P^t_{train}$ for the current task. On the other hand, the old classification head $CL_{t-1}$ can be used to represent the old knowledge of the model. Samples from support set $K_t$ can be passed through a list of classification head module $CL_{i}$ from the first task to the latest old task $t-1$ (\textit{i.e} $i
\in [0, t-1]$) in order to obtain prediction logits.\\
\textbf{Cross-Distillation Loss Function} is common in most of exemplar-based methods. This is constructed by combining cross-entropy, and distillation loss function. The cross-entropy loss function helps minimize the overall empirical errors when learning on new category set $C_t$ at task $t$. Meanwhile, the distillation loss function plays a role in distilling the old model $M_{t-1}$ from previous tasks into the current model $M_t$ to avoid catastrophic forgetting. Let's consider the incremental learning model at a specific learning stage $t$ where it has obtained $t-1$ numbers of classification heads. New classification head $CL_t$, which is now added to learn on new task $t$, produce the prediction logits as $o(x) = [o_1(x), o_2(x), ..., o_t(x)]$ for any input $x$. Similarly, output logits which are produced by old classification head can be represented as $\hat{o}(x) = [\hat{o}_1(x), \hat{o}_2(x), ..., \hat{o}_{t-1}(x)]$. With these representation, the distillation loss can be computed for all samples from exemplar sets $K_{t-1}$ and from new classes $P^t_{train}$ (\textit{i.e.} $K_{t-1} \cup P^t_{train} = V_{train}^t$) as follows: 

\begin{equation}
\begin{aligned}
L_{d} &=\sum_{x \in V_{train}^t} \sum_{k=1}^{t-1}-\hat{\pi}_{k}(x) \log \left[\pi_{k}(x)\right], \\
\hat{\pi}_{k}(x) &=\frac{e^{\hat{o}_{k}(x) / T}}{\sum_{j=1}^{t-1} e^{\hat{o}_{j}(x) / T}}, \quad \pi_{k}(x)=\frac{e^{o_{k}(x) / T}}{\sum_{j=1}^{t-1} e^{o_{j}(x) / T}},
\end{aligned}
\label{eq:distillation_loss}
\end{equation}

where T plays as the temperature scaling factor. 
Meanwhile, the groundtruth label for each sample $x$ (\textit{i.e.} y(x)) for new category sets along with softmax of logits of the $k$-th category set (i.e. $p_k=softmax(o_k(x))$ ) can be used to compute the cross-entropy loss function as follows

\begin{equation}
    L_{c}=\sum_{x \in V_{train}^t} \sum_{k=1}^{t}-y(x) \log \left[p_{k}(x)\right],
\label{eq:cross_entropy}
\end{equation}

The final cross-distilled loss function; therefore, can be obtained by combining distillation loss function in Eqn.\ref{eq:distillation_loss} and cross-entropy loss function in Eqn.\ref{eq:cross_entropy}

\begin{equation}
    L = \alpha L_d + (1-\alpha) L_c,
\end{equation}

where the scalar value $\alpha$ controls the balance between the two functions.
\section{Methodology}
The majority of current pill identification methods rely on RGB images. Therefore, to the best of our knowledge, most existing systems fail to address hard examples (\textit{e.g}, pills with very similar shapes and colors)~\cite{9157392}. This problem becomes more challenging in the context of the class incremental learning. In this problem, we have to cope with two issues at the same time: 1) recognizing pill instances that belong to the novel classes, and 2) not forgetting the previously learned knowledge of the old ones.
We seek in this study robust domain-specific knowledge, which could be in good companion to traditional RGB image stream. However, the introduction of an additional stream issue a different challenge; the significant need for a stream integration method . To tackle such a challenge, we propose an Incremental Multi-stream Intermediate Fusion framework (IMIF). The IMIF allows additional information streams to be effectively propagated during the incremental learning phase. In the following subsections, we briefly define the multi-stream class incremental learning method and describe how it can be decomposed into different components.

\subsection{Multi-stream Class Incremental Learning Model}

We define a multi-stream class incremental learning model \textbf{M} as a combination of three key components: 1) a single stream base method \textbf{X}, 2) an additional stream of information \textbf{Y}, and 3) a method of fusing stream \textbf{Z}.
\begin{center}
    \textbf{M} = Base method \textbf{X} + Feature stream \textbf{Y} + Fusion mechanism \textbf{Z}
\end{center}
At this point, the base method represents any method that follows the general setting described in Sec.\ref{subsec:conventionalCIL}. \textbf{Y} serves as a piece of additional domain information that gives cues to the learning model apart from RGB images. 
Normally, \textbf{Y} is specific to the domain of the task. Lastly, \textbf{Z} presents a fusion mechanism that enables method \textbf{M} to incorporate additional information stream \textbf{Y} into the incremental learning process. From this decomposition, our CG-IMIF replaces: 1) the representative stream \textbf{Y} with color-specific information, and 2) the fusion technique \textbf{Z} with the proposed IMIF.
In the following, we describe our proposed Color histogram guidance stream and Incremental Multi-stream Intermediate Fusion technique in Sec.\ref{ssec:domain-specific information extractor} and Sec.\ref{subsec:fusion}, respectively.

% \subsection{Base Method X}

% At this point, base method X can be represented with any exemplar-based methods which follows the general architecture described in Sec.\ref{subsec:conventionalCIL}.

% Baseline methods are those existing works that have been proposed to solve the traditional \textcolor{blue}{\st{Class Incremental Learning}} \textcolor{red}{CIL} problem. \textcolor{red}{In the task of the prescription pill images}, we inherited various exemplar-based class incremental learning approaches to study the effect of \textcolor{blue}{\st{our}} \textcolor{red}{[Please do not use too much "our"]} the proposed framework IMIF when being integrated into the original one. In general, most of the exemplar-based CIL methods \textcolor{blue}{\st{would follow to}} consist of two following components:

\begin{figure}[t]
    \centering
    \includegraphics[width=0.5\textwidth]{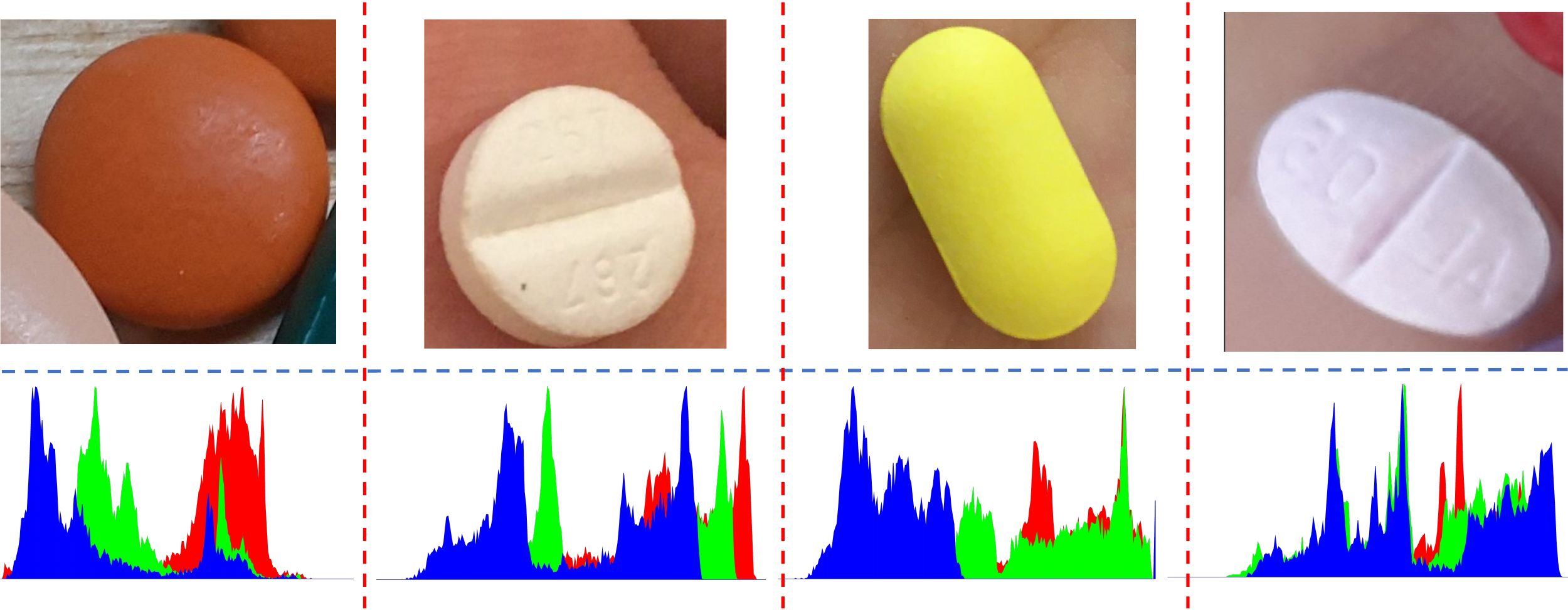}
    \caption{Samples of pill images from VAIPE-PCIL dataset are shown on the first row. The second row exhibits the corresponding color histogram information for images on the upper row respectively.}
    \label{fig:example_color_histogram}
\end{figure}

\subsection{Color Histogram Guidance Stream} \label{ssec:domain-specific information extractor}

Pill images compose of various features which can be used as discriminative factors in classification problems. Among those is color distribution information which can be approximated by the color histogram of pill images. The color histogram represents the color distribution of the input image in terms of three different channels: red, green, and blue. In detail, it is often encoded into a single vector where the indexes of entries are mapped to a set of all possible colors. In grey-scale images, values of vector entries store the frequency that counts the total number of pixels having the intensity (\textit{i.e.}, color value), which is hashed by the corresponding index of the vector. However, in a three-dimensional image, color ranges for each channel are associated across color channels to formulate a unique combination. This combination accounts for those pixels of which color ranges lie in three discrete ranges of value: $[r_i, r_j];[g_k, g_l];[b_m, b_n]$. The color feature vectors are useful to make a distinction among pill instances that have similar shapes but different colors. The color ranges for each channel of the RGB images are divided into eight segments in our problem, where each segment represents 32 different consecutive color values. After that, the color histogram vector can be obtained by accumulating the quantities of pixels assigned to a specific color range for each channel to result in a vector with 8$\times$8$\times$8 = 512 elements. Fig.~\ref{fig:example_color_histogram} illustrates a few examples of extracting the color histogram stream for each corresponding cropped pill image.

\begin{figure}[ht]
    \centering
    \includegraphics[width=0.75\textwidth]{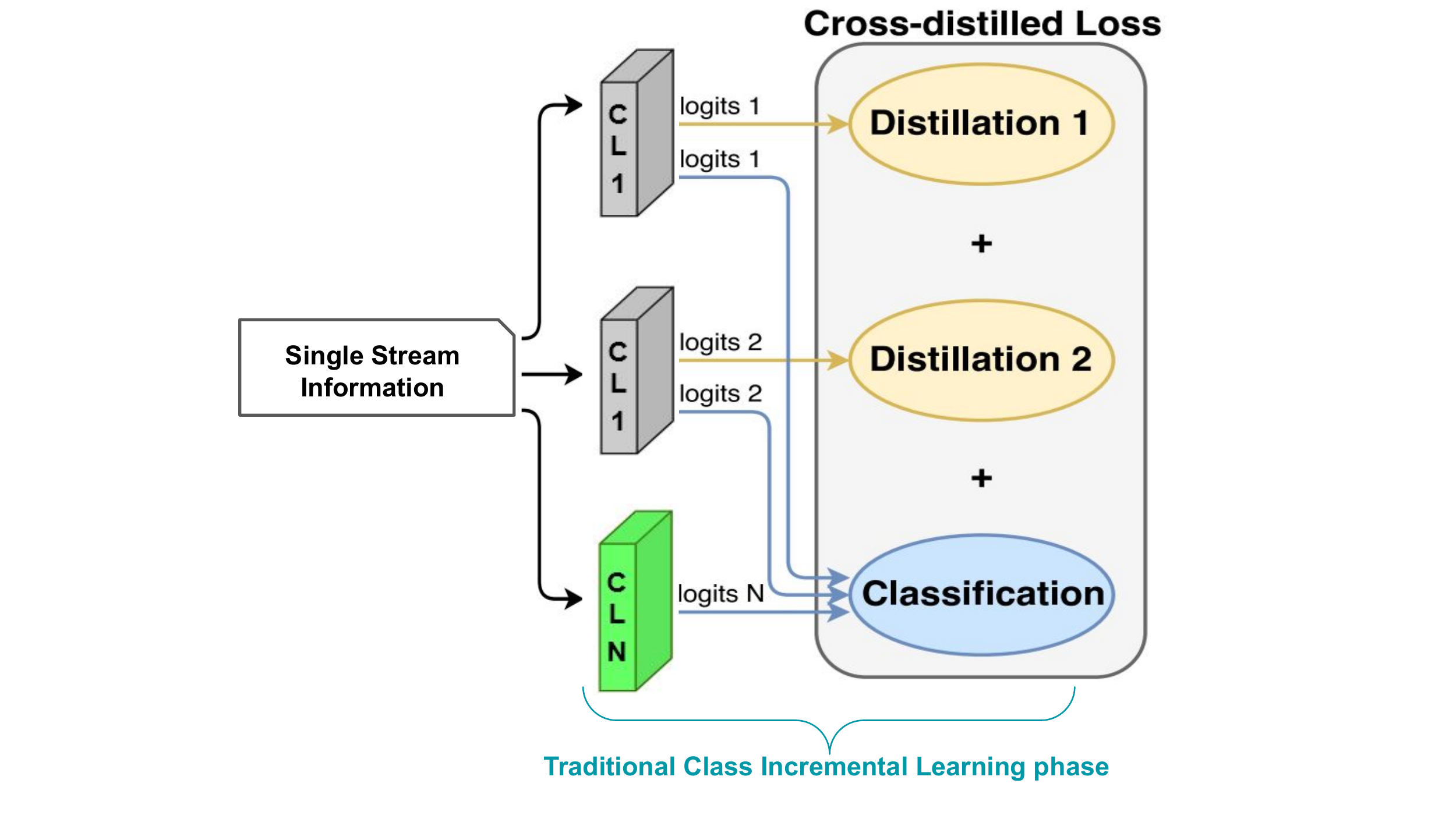}
    \caption{The traditional training paradigm with only single-stream information used by almost existing exemplar-based CIL methods.}
    \label{fig:traditional_single_stream}
\end{figure}

% \begin{figure*}[ht!]
% 	\centering
% 	{
%     \includegraphics[width=0.75\textwidth]{images/traditional_single_stream.pdf}
% 		\caption{The traditional training paradigm with only single-stream information use by almost existing exemplar-based CIL methods.}
% 		\label{fig:traditional_single_stream}}
% 	\centering
% 	{
%     \includegraphics[width=0.75\textwidth]{images/X-CG-IMIF.pdf}
% 		\caption{Our proposed CG-IMIF architecture composes of: 1) color histogram feature extraction (orange block), and 2) intermediate fusion framework (purple block) to incorporate additional information stream.}
% 		\label{fig:our_X_CG_IMIF}}
% \end{figure*}

\subsection{Incremental Multi-stream Intermediate Fusion Technique}
\label{subsec:fusion}

\textbf{Traditional Early Fusion}
A naive fusion technique \textbf{Z} concatenates different streams of information right after the feature extraction phase. Specifically, feature vectors $f_{r} \in \mathbb{R}^{d_r}$, and $f_Y \in \mathbb{R}^{d_Y}$ are extracted from raw RGB images, and additional information stream \textbf{Y}, respectively. Both of these features are then fed into separate projection layers to project into the same latent space. In practice, the projection layers are implemented by a single hidden layer controlled by parameters $\Theta^p=[W^p, b^p]$ as follows
\begin{equation}
s_{r} = \sigma(W^p_r.f_r+b^p_r),\\  s_{Y} =\sigma(W^p_Y.f_Y+b^p_Y), \\  s_{g} = [s_r, s_Y].
\label{eq:projection_layer}
\end{equation}
The projected vectors are then concatenated to obtain the global feature vector $s_g \in \mathbb{R}^{d_g}$. This global feature $s_g$ can be considered as a single input into any traditional single stream CIL methods as shown in Fig.~\ref{fig:traditional_single_stream}.

\begin{figure}[htbp]
    \centering
    \includegraphics[width=0.75\textwidth]{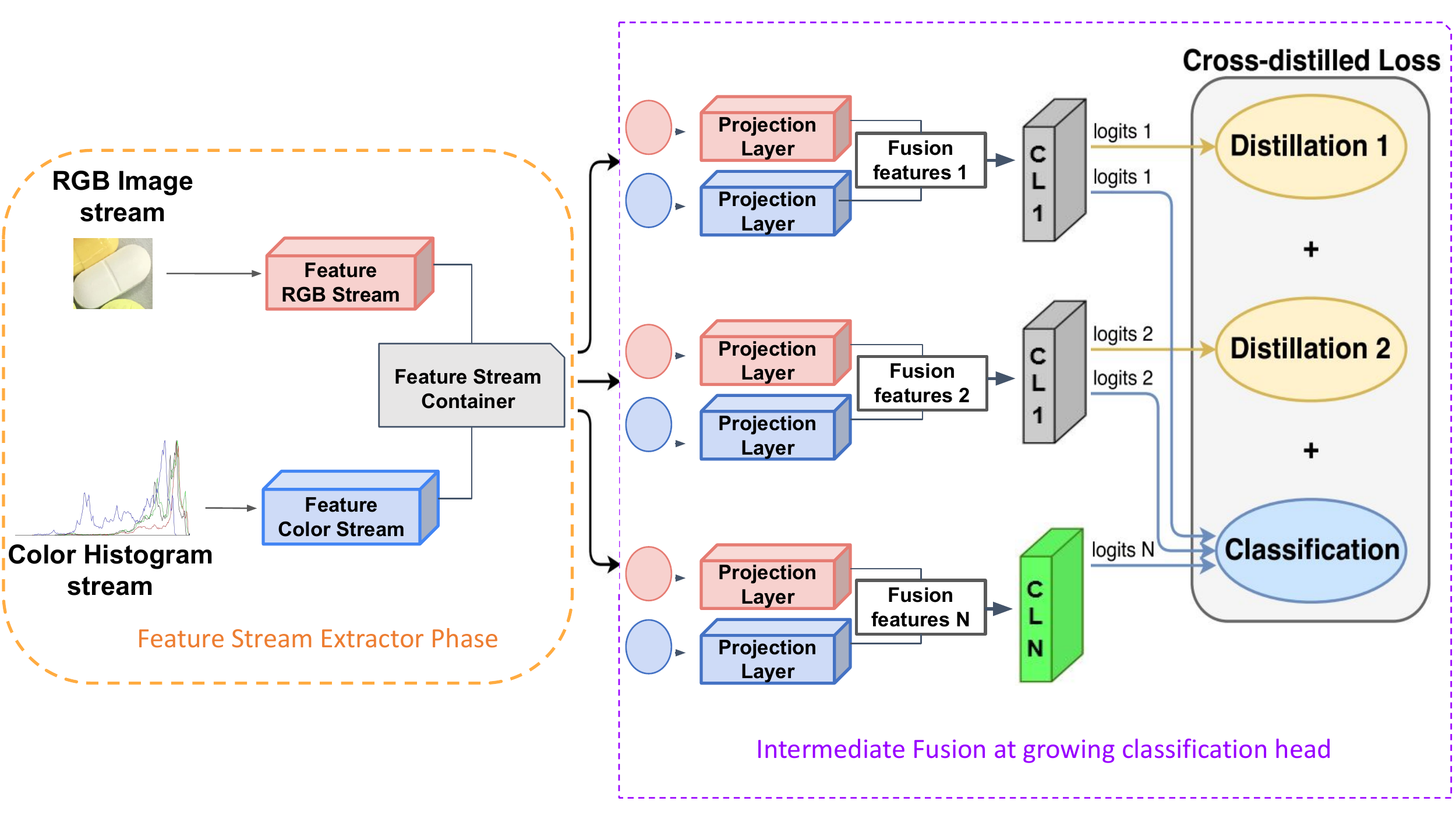}
    \caption{Our proposed CG-IMIF architecture composes of: 1) color histogram feature extraction (orange block), and 2) intermediate fusion framework (purple block) to incorporate additional information stream.}
    \label{fig:our_X_CG_IMIF}
\end{figure}
\noindent\textbf{Intermediate Fusion}
We observed that fusing information stream in an early manner for a class incremental learning problem is not optimal. The global feature $s_g$ is problematic since it is regularly updated at each incremental task. As a result, the projection layer in the early phase can not find good parameters that balance the performance of old and novel categories in different tasks. Therefore, we propose to relocate the projection layer to the intermediate phase instead (\textit{i.e}, incremental learning phase) by initiating an entirely novel projection layer in an incremental manner. When $C^t$ from new task $t$ arrives at the system, a new classification layer $CL_t$ is created . This layer accompanies by an attached projection layer specific to the information stream of a specific task. Therefore, the parameters controlling the projection layer for each information stream are different from those defined in the early fusion.
\begin{equation} 
\Theta^{p^t}_{r} = [W^{p^t}_r, b^{p^t}_r],
\Theta^{p^t}_{Y} = [W^{p^t}_Y, b^{p^t}_Y].
\end{equation}

% In this way, the projection layer could maintain its optimal performance for old tasks without significantly adjusting its weight toward the novel ones. The intermediate fusion behavior is illustrated in Fig.~\ref{fig:our_X_CG_IMIF}.
\section{Experiments}
\subsection{Dataset}

We employ a real-world image dataset, namely VAIPE-Pill (VAIPE Pill Identification) \cite{VAIPE} to exploit CIL capability on the pill image classification problem. This dataset is created to promote the research on recognizing distinct types of medicines from mobile devices. The dataset contains 7,294 pill images of 262 categories taken in real-world scenarios. The characteristics of VAIPE-Pill dataset are illustrated in Tab.~\ref{tab:data_stat}.

To facilitate research of CIL in pill image classification tasks, we derive a dataset version, namely VAIPE-PCIL (VAIPE Pill Class Incremental Learning) dataset from the original VAIPE-Pill data. VAIPE-PCIL is obtained by cropping pill instances from the original data. We only select those categories which satisfy either of the following conditions: 1) the number of samples should not be too small (\textit{i.e.}, and 2) larger than 10 samples), image size of samples should be at least $64 \times 64$. Samples of pill image from VAIPE-PCIL can be found in Fig.~\ref{fig:overview}. All of our experiments are conducted on the VAIPE-PCIL dataset to study the performance of CG-IMIF. 

\begin{table}
    \centering
    \caption{Statistics of VAIPE-Pill dataset on different characteristics.}
    {
\begin{tabular}{c|c|c|c}
\hline \textbf{Characteristic} & \textbf{Training set} & \textbf{Testing set} & \textbf{Total} \\
\hline Number of images & 6,461 & 833 & 7,294 \\
Number of pill categories & 262 & 262 & 262 \\
Instances per category & $179.75$ & $23.56$ & $203.2$ \\
Image size (pixel$\times$pixel, mean) & $3,311 \times 3,276$ & $3,276 \times 3,469$ & $3,300 \times 3,400$ \\
Instances per image & $7.28$ & $7.4$ & $7.3$ \\
Number of bounding box annotations & 47,097 & 6,174 & 53,271 \\
Number of categories per image & $5.18$ & $5.76$ & $5.32$\\
\hline
\end{tabular} }
\label{tab:data_stat}
\end{table}

\subsection{Experimental Protocol}

\textbf{Settings} We follow the standard benchmark protocol proposed in \cite{Hou_2019_CVPR}. We fix class arrangements in random order. 
After each training stage $t$, the resulting learner is evaluated on the testing data $\bigcup_{i=1}^{t}P_{test}^t$ which represents for all of the testing data up to the current task $t$. Since no test data from the previous learning stage are hidden from the learner, it is guaranteed that no overfitting can occur. 

There are two commonly different task evaluation settings in class incremental learning: task-awareness and task-agnostic. The first setting is much easier for the algorithm since it has access to the task ID (\textit{i.e.}, ID or set of categories) about the incoming test data. Therefore, it is reasonable to only use the corresponding classification head in the incremental learning phase, which is trained on that task-ID to evaluate the performance. This task setting, however, is not practical in many real-world circumstances since task-ID is not always available. We evaluate our performance in terms of task-agnostic instead. In task-agnostic, the model is not given the task identities of the test data. Hence, the evaluation results are achieved by taking the results of all prediction logits, which are predicted by all of the classification heads. In this way, the model has to learn to resolve the confusion among classes from a different set of classes.
 
 \noindent\textbf{Evaluation Metrics} We adopt two commonly used benchmark metrics from \cite{Hou_2019_CVPR} for CIL problems: average accuracy and average forgetting rate. The average accuracy and forgetting rate records of performance for each incremental learning phase are often if a single number is preferable. Meanwhile, the average phase accuracy and forgetting rate would be used to observe learning behaviors during incremental tasks for each method.
\subsection{Implementation Details}

All of our experiments and methods are implemented with Pytorch \cite{NEURIPS2019_9015} and trained on a single NVIDIA GeForce RTX 3090. We inherit the codebase from FACIL \cite{masana2020class}. They have already implemented various state-of-the-art methods for CIL problems in a well-structured manner. Details of base models as well as the implementation of our IMIF framework are discussed below. In all experiments, we attach our IMIF framework to several state-of-the-art methods in CIL: BiC\cite{DBLP:journals/corr/abs-1905-13260}, EEIL\cite{DBLP:journals/corr/abs-1807-09536}, and LUCIR\cite{Hou_2019_CVPR}. Since these methods followed the common prototype of exemplar-based methods, we discussed some of the general settings of exemplar-based methods in our experiments before diving into details about the setting of each base one. There are two common strategies to reserve samples for old classes: 1) exemplar-management stores a constant number of samples for each old class, or 2) it maintains a fixed capacity (\textit{e.g}, $R_{total} = 2,000$ for CIFAR-100 \cite{cifar} and $R_{total} = 20,000$ for ImageNet \cite{deng2009imagenet}). In our experiments, we follow the first setting since it is usually more challenged than the second one. In addition, the exemplars are randomly selected among different categories. For the training network, we made use of 50-layer ResNet \cite{DBLP:journals/corr/HeZRS15} with no pre-trained weights as the backbone for feature extraction module, which is applicable to the VAIPE-PCIL dataset. To fairly compare the performance, we fixed the number of training epochs (200 epochs) across different methods. The learning rate is initialized with 0.1 and is divided by 1.5 if the loss function suffers from non-decreasing circumstances for a specific number of attempts (\textit{e.g}, $lr_{patience} = 5$). The networks are trained using stochastic gradient descent with mini-batches of 32 samples. The training images are resized to the same shape of $256 \times 256 \times 3$ with only one transformation (\textit{e.g}, flipping). The class orders across different methods are randomly fixed for a fair comparison.

In terms of configuration for base methods, we follow the same settings for the original version BiC \cite{DBLP:journals/corr/abs-1905-13260}, and our improved version BiC-CG-IMIF. BiC\cite{DBLP:journals/corr/abs-1905-13260} proposed to integrate a bias correction layer attached to the end of each classification head to adjust the classification score. The number of training epochs for the bias correction layer is 200 epochs in our setting. Moreover, we set 0.1 as the ratio of the number of exemplars that are used for the validation. EEIL \cite{DBLP:journals/corr/abs-1807-09536} performs an additional fine-tuning phase after each official training phase to balance the performance between old and novel categories. In our experiments, we fix 40 as the number of epochs for fine-tuning and the learning rate fine-tuning factor as 0.01 across different methods. We also adhere to the base setting of LUCIR\cite{Hou_2019_CVPR} method where they removed ReLU in the penultimate layer to take both positive and negative values. 
For the IMIF framework, the projection layer implemented is represented by a single hidden layer. Therefore, the output size for different projection layers should be the same so that the transformed feature vectors, then can be fused in the shared space. In terms of the color-guided information, color ranges for each channel of the RGB images are divided into 8 segments where each segment represents 32 different consecutive pixel values.

\subsection{Experimental Results}
\label{subsec:exp_result}
\begin{figure*}[htbp]
	\centering
	{
    \includegraphics[width=\textwidth]{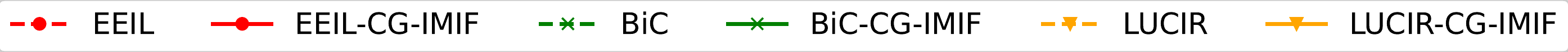}
		\begin{tabular}[width=\textwidth]{ccc}
			\subfigure[5 tasks] {\includegraphics[height=25mm]{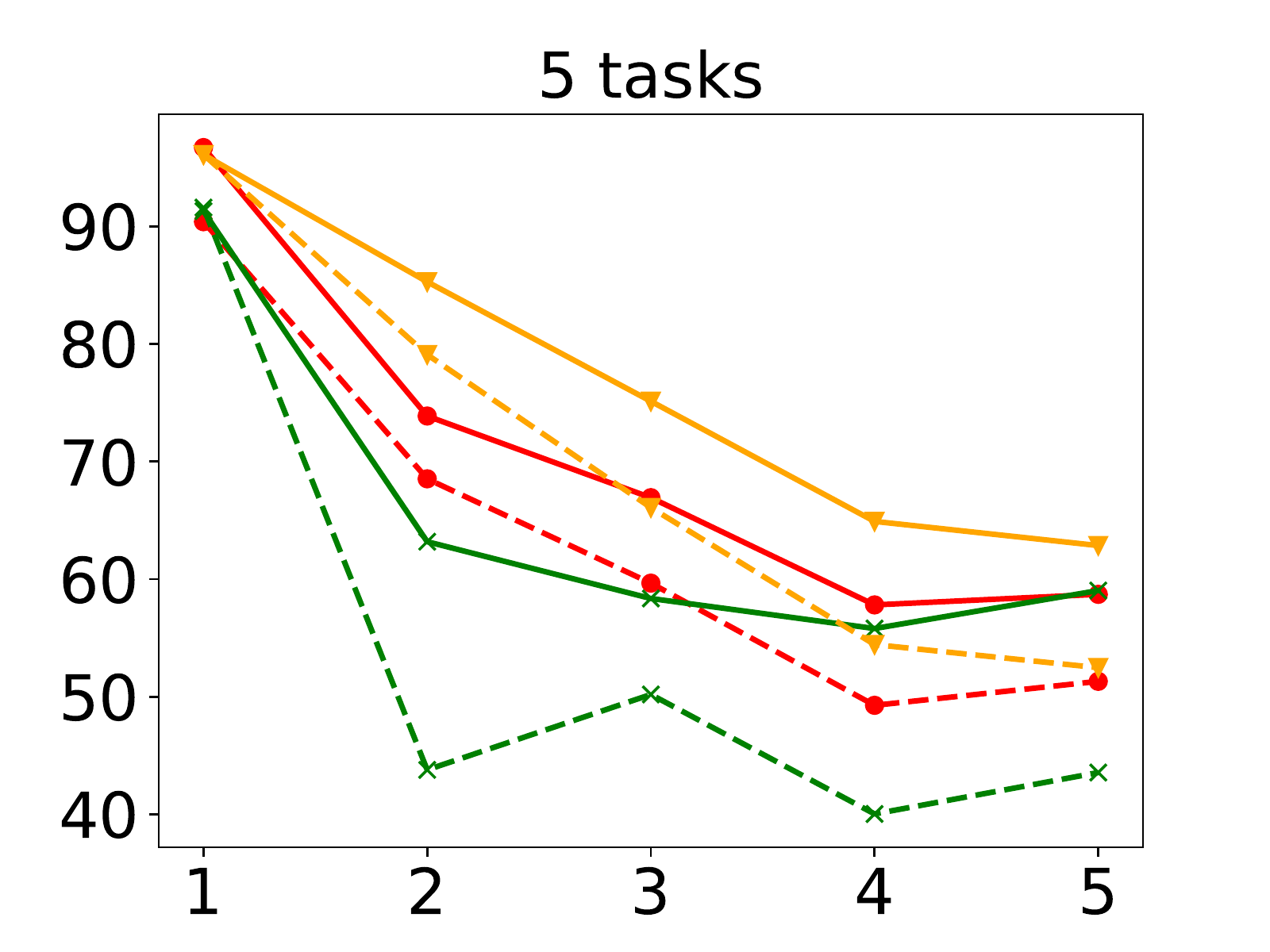}}  &
			\subfigure[10 tasks] {\includegraphics[height=25mm]{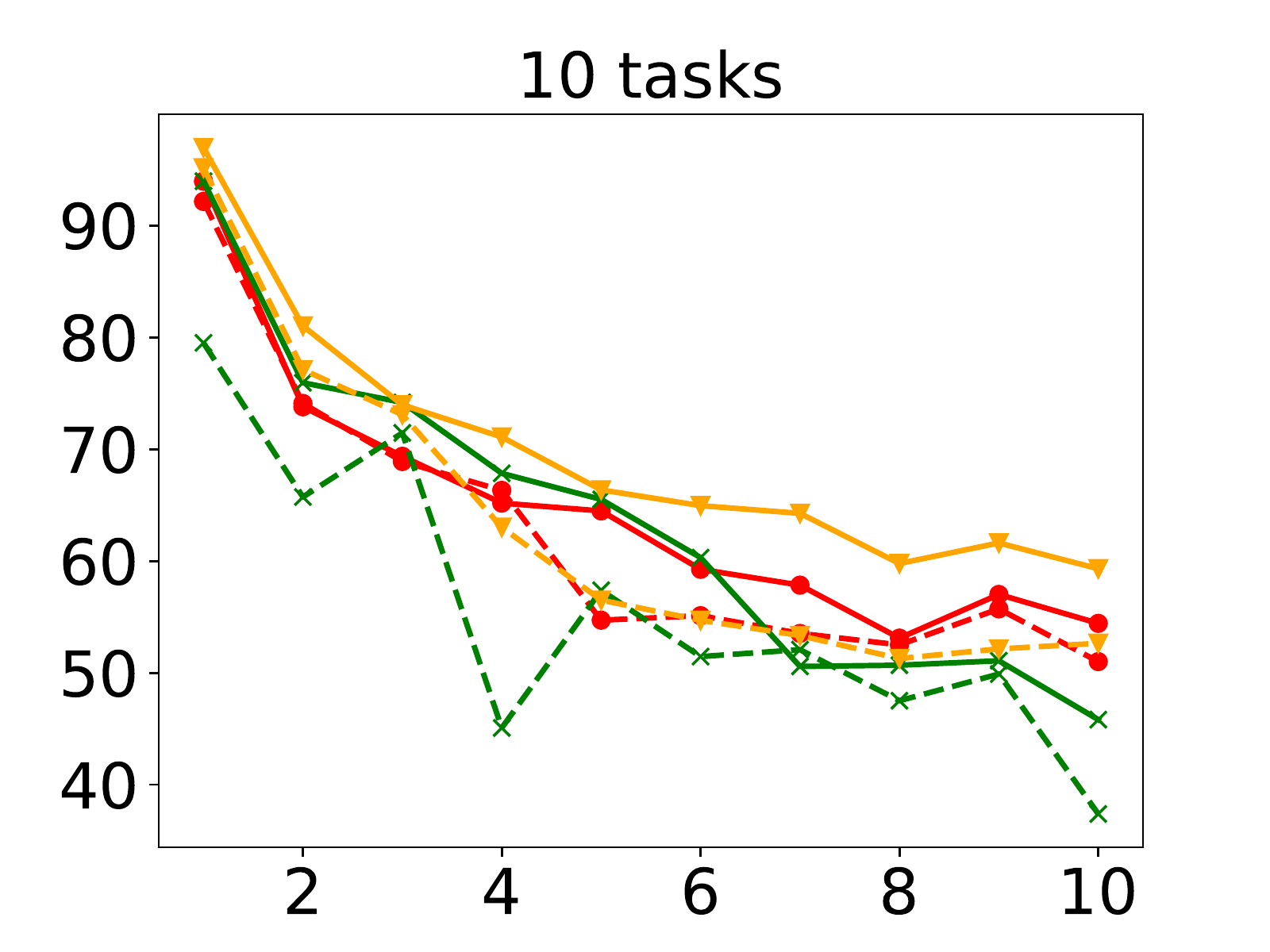}}   &
			\subfigure[15 tasks] {\includegraphics[height=25mm]{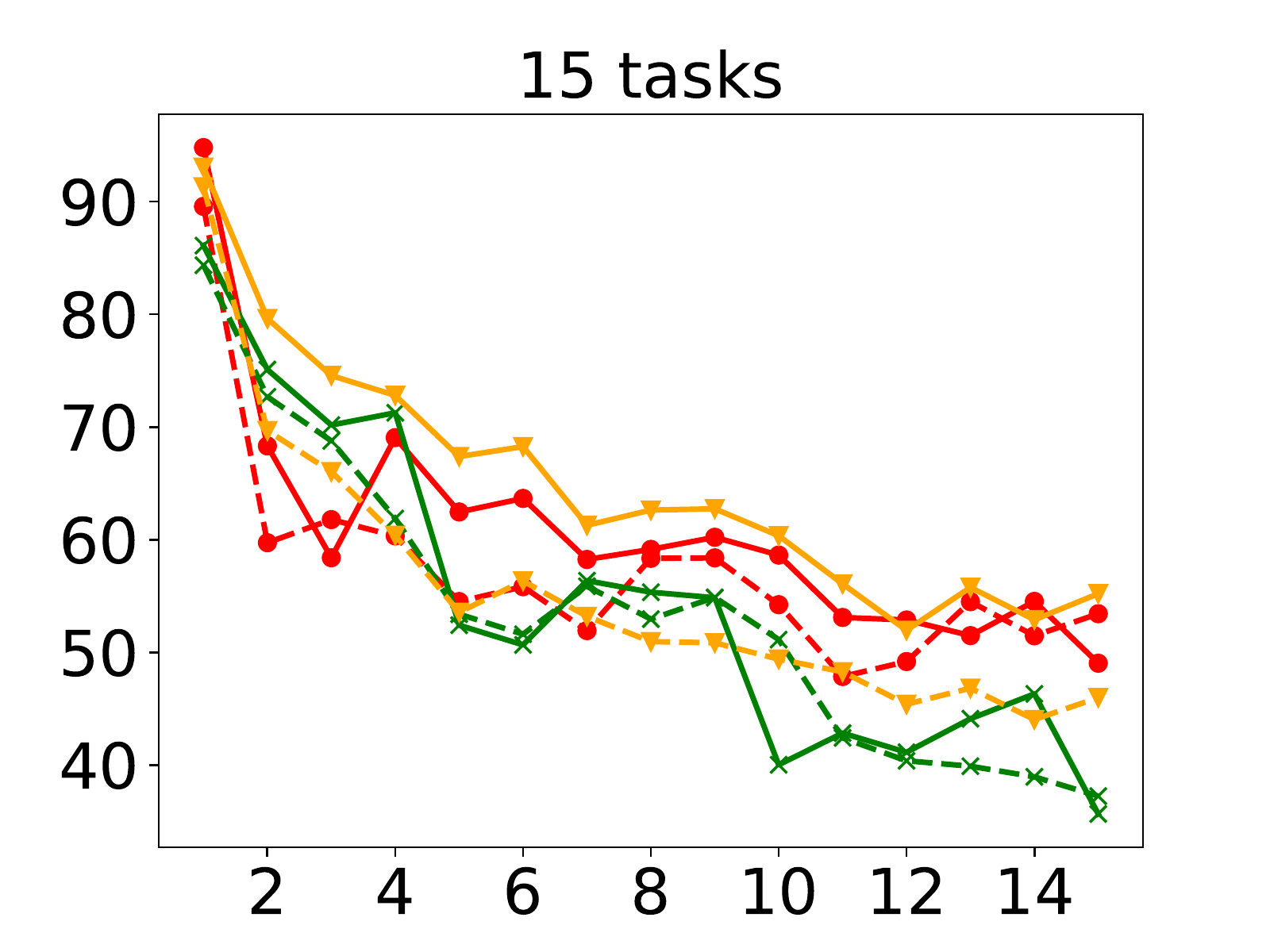}}
		\end{tabular}
		\caption{\textbf{Incremental accuracy} for different task settings among the original version and our method CG-IMIF.}
		\label{fig:average_phase_acc}
	}
% 	\end{figure*}
% 	\begin{figure*}[t]
	\centering
	{
		\begin{tabular}[width=\textwidth]{cccc}
			\subfigure[5 tasks] {\includegraphics[height=25mm]{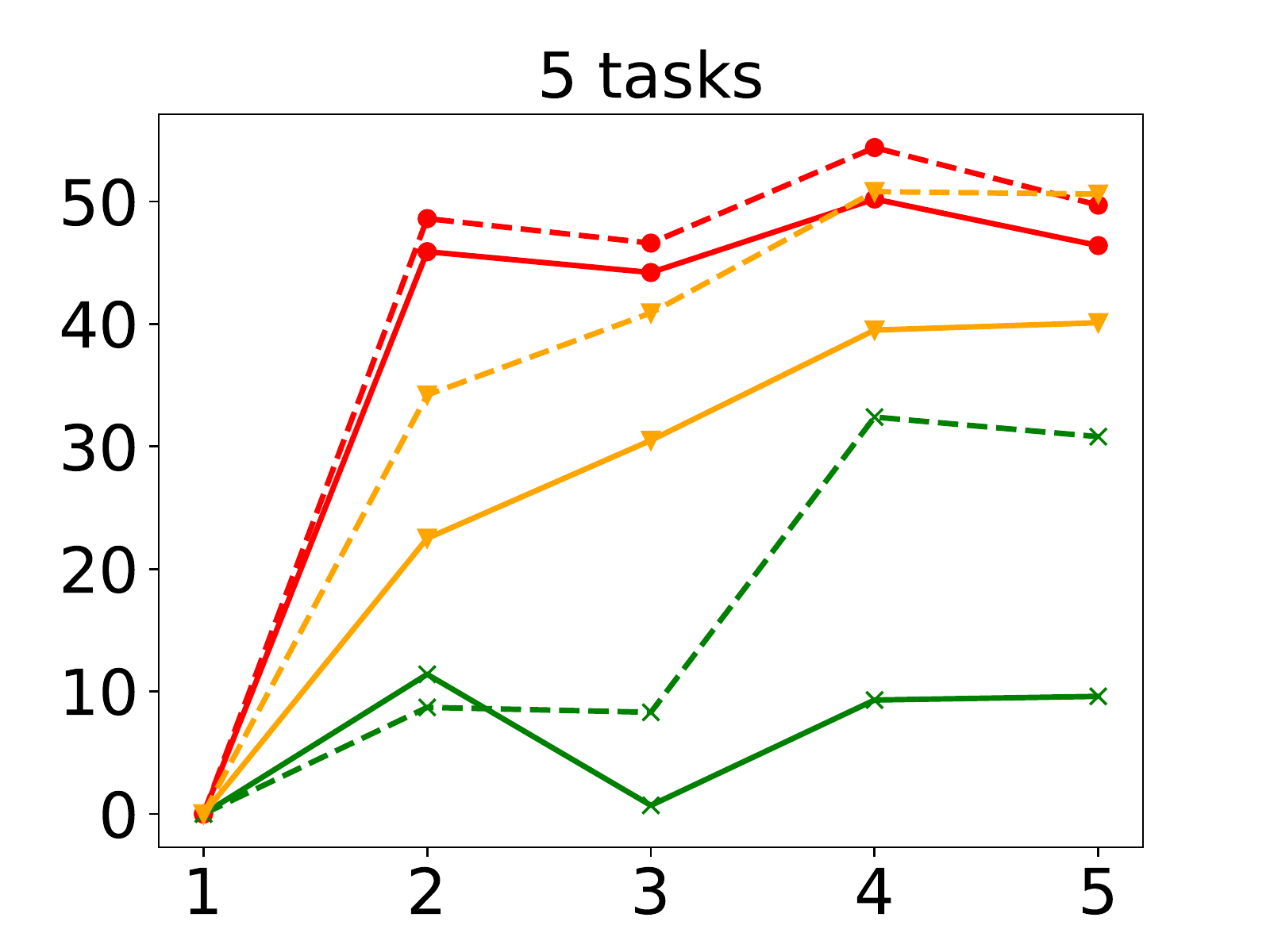}}  & 
			\subfigure[10 tasks] {\includegraphics[height=25mm]{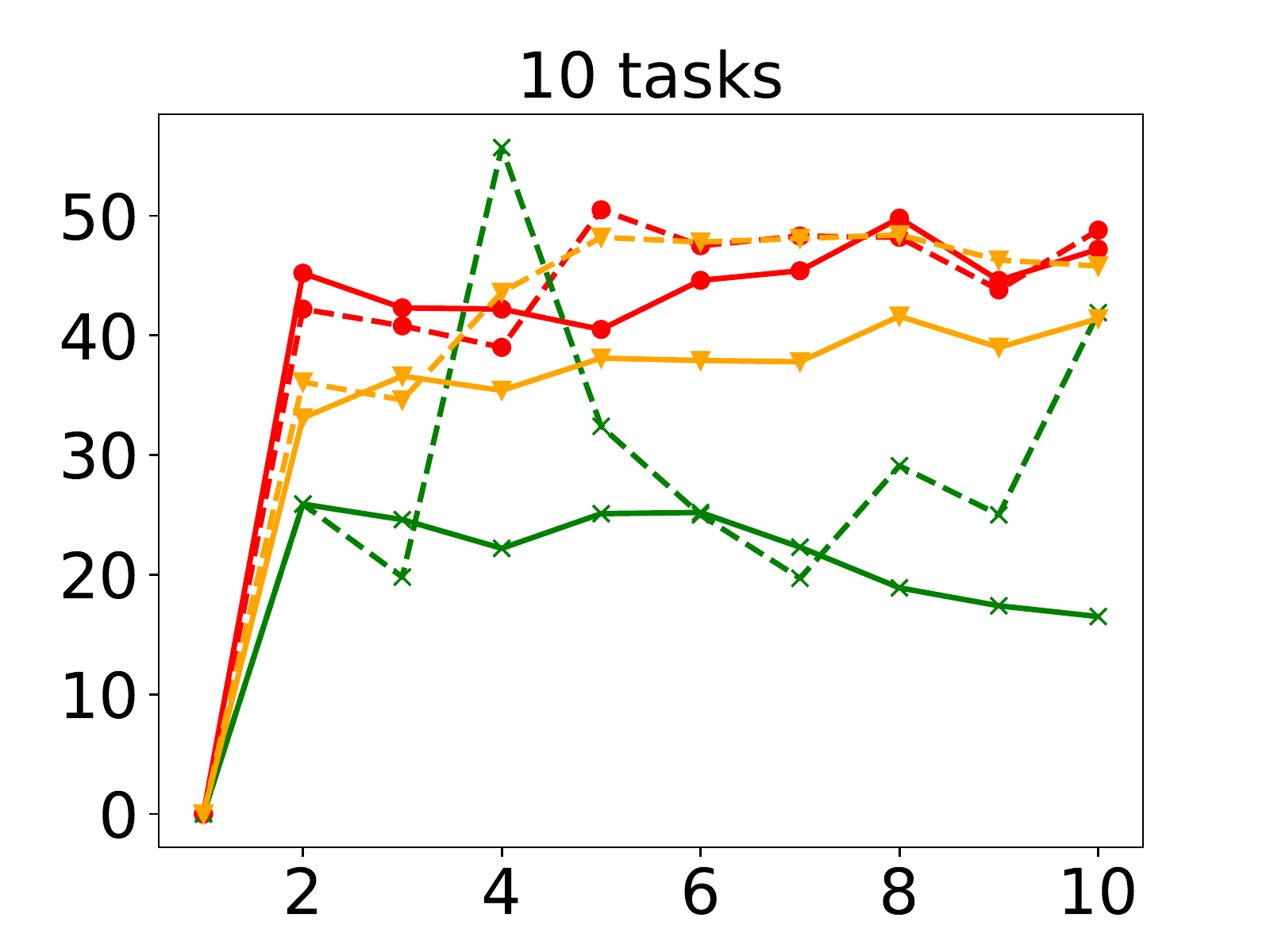}}   &
			\subfigure[15 tasks] {\includegraphics[height=25mm]{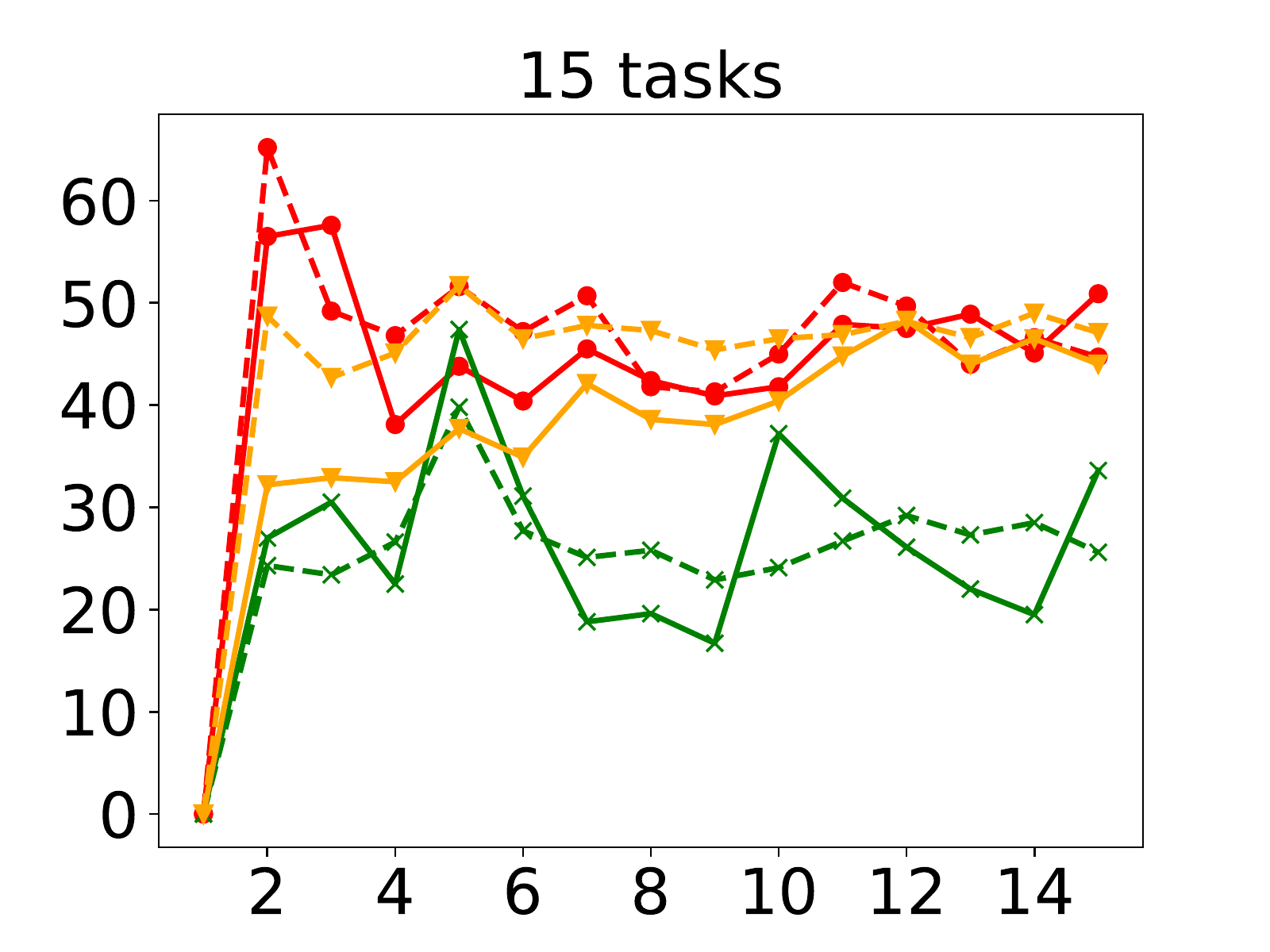}} 
		\end{tabular}
		\caption{\textbf{Incremental forgetting rate} for different task settings among the original version and our method CG-IMIF.}
		\label{fig:average_forgetting_rate}
	}
\end{figure*}

\begin{table*}[htbp]
  \small
  \centering
  \caption{
  Average accuracy $\bar{\mathcal{A}}$ (\%) and forgetting rate $\mathcal{F}$ (\%) of CG-IMIF compared to other state-of-the-art results in different task settings. Best scores are marked in bold for both evaluation metrics.
  }
  {
  \begin{tabular}{llccccccccccccccc}
  \toprule
  \multirow{2.5}{*}{Metric} & \multirow{2.5}{*}{Method} & \multicolumn{3}{c}{\emph{Task Settings}} \\
  \cmidrule{3-5}
  & & $N$=5 & $N$=10  & $N$=15 \\ %\emph{w/} ours
   \midrule
  & EEIL\cite{DBLP:journals/corr/abs-1807-09536} & 63.83 & 62.40 & 57.41 &\\
  & \textbf{EEIL-CG-IMIF} & \textbf{\textbf{70.80}} & \textbf{\textbf{64.85}} & \textbf{\textbf{60.93}} \\
  \cmidrule{2-13}
  \textbf{\emph{Average acc.} (\%) $\uparrow$} & BiC \cite{DBLP:journals/corr/abs-1905-13260} & 53.83 & 55.75 & 53.77  \\
  {$\bar{\mathcal{A}}=\frac{1}{n}\sum_{i=1}^{n}\mathcal{A}_i$}& \textbf{BiC-CG-IMIF} & \textbf{\textbf{65.53}} & \textbf{\textbf{63.59}} & \textbf{\textbf{54.83}}\\
  \cmidrule{2-13}
    \multirow{1}{*} & LUCIR \cite{Hou_2019_CVPR}  & 69.63 & 62.90 & 55.49\\
  & \textbf{LUCIR-CG-IMIF} & {\textbf{76.85}} & {\textbf{69.94}} & {\textbf{64.97}} \\
  \midrule
   & EEIL \cite{DBLP:journals/corr/abs-1807-09536} & 49.82 & 45.46 & 48.27 \\
  & \textbf{EEIL-CG-IMIF } & \textbf{\textbf{46.68}} & \textbf{\textbf{44.64}} & \textbf{\textbf{46.23}} \\
  \cmidrule{2-13}
  \textbf{\emph{Forgetting rate}. (\%) $\downarrow$} & BiC \cite{DBLP:journals/corr/abs-1905-13260} & 20.05 & 30.50 & 26.93  \\
  \multirow{1}{*}{$\bar{\mathcal{F}}=\frac{1}{n}\sum_{i=1}^{n}\mathcal{F}_i$} & \textbf{BiC-CG-IMIF } & \textbf{\textbf{7.75}} & \textbf{\textbf{22.01}} & \textbf{27.35} \\
  \cmidrule{2-13}
    \multirow{1}{*} & LUCIR \cite{Hou_2019_CVPR}  & 44.13 & 44.32 & 47.11\\
  & \textbf{LUCIR-CG-IMIF } & {\textbf{33.15}} & {\textbf{37.88}} &  {\textbf{39.79}} \\
  \bottomrule
    \multicolumn{13}{l}{${}^{\diamond}$ Using the similar exemplar settings and selection for fair comparison.}
\end{tabular}
}
  % \cotronlcaptionvsapce
  \label{tab:experimental_result}
  % \cotronlvsapce
\end{table*}
We evaluate our proposed CG-IMIF approach and report the overall performance in comparison with several state-of-the-art approaches in Tab.\ref{tab:experimental_result} Experimental results show that most of the state-of-the-art approaches attached with our proposed IMIF tool and color-specific information as additional stream help to achieve consistent improvements over task settings. 
The setting consists of three tasks in total where the number of categories is uniformly distributed for 5, 10, and 15 tasks. It is noticeable that the lower score of forgetting rate indicates that the model is more unlikely to forget about old knowledge. In addition, average phase accuracy and forgetting rate is also illustrated in Fig.~\ref{fig:average_phase_acc},\ref{fig:average_forgetting_rate} to inspect the learning behaviour of each method through incremental phases. Dashed and solid lines with different colors are utilized to differentiate the base ones (X) and our CG-IMIF, respectively. In terms of the average phase accuracy, LUCIR-CG-IMIF obtains the highest performance where it can consistently and significantly surpass other methods (also the base one-LUCIR). On the other hand, BiC-CG-IMIF is better at mitigating the forgetting constraint. However, it is not consistent over tasks and the curve fluctuates. One possible explanation is that the bias layer inside the traditional BiC method and BiC-CG-IMIF might cause the model to sacrifice the performance of the current task to maintain the memory of the old ones. 
\section{Ablation Studies}
To examine the effect of additional information stream usage and fusion framework, we perform extensive ablation studies. This is aim to observe the effect of different components in our proposed framework where LUCIR is chosen as the base method. LUCIR is preferable because of the consistency and high performance of LUCIR across task settings in the experimental results which have been discussed in Sec. \ref{subsec:exp_result}.
% \subsection{Importance of Information Stream}
In addition to color information, edge signals might be a good candidate to discriminate different pill categories based on their shape. To understand the importance of different stream usage in our method, we compare 4 different settings: 1) RGB image only, 2) RGB and edge images, 3) RGB and color histogram, and 4) a combination of all three streams. Each separated row in Tab.\ref{tab:ablation_study}  refers to each scenario of information stream usage with two different fusion techniques. Concretely, the setting that combines RGB and color histogram streams achieves the highest score. One possible explanation for this result is that the edge signal might not be sufficiently strong to push the performance. 

In addition, we implement the basic fusion technique where additional information streams are fused in an early manner. Each separated row in Tab.\ref{tab:ablation_study} illustrates the results of two different fusion mechanisms. Our fusion technique (on the second line of each row) outperforms the traditional one in various metrics and task settings. The best result is LUCIR-CG-IMIF which integrates color histogram information into the traditional LUCIR method with IMIF.

% \subsection{Effectiveness of IMIF Framework}
% In addition, we implement the basic fusion technique where additional information streams are fused in an early manner. Each separated row in Tab.\ref{tab:ablation_study} illustrates the results of two different fusion mechanisms. Our fusion technique (on the second line of each row) outperforms the traditional one in various metrics and task settings. The best result is LUCIR-CG-IMIF which integrates color histogram information into the traditional LUCIR method with IMIF.

% The intermediate fusion can help the base model attached with any additional stream of information achieve better results than early fusion. In order to prove this statement, we perform a fair comparison between early fusion and intermediate fusion techniques across different streams of information: color stream, and edge image stream. We implemented the projection layer for each fusion mechanism in exactly the same way by using a single hidden layer.

% show a table with early fusion, intermediate fusion for the edge, color, and triple stream
\section{Discussions}
\begin{table*}
  \small
  \centering
  \caption{
  Ablation performance to compare variants of combination which utilize different information streams as well as different fusion techniques. The Combination which achieves highest performance over different tasks is our CG-IMIF and is marked in green.
  }
  {
  \begin{tabular}{llccccccccccccc}
  \toprule
  \multirow{2.5}{*}{Variant of Combination} & \multicolumn{3}{c}{\emph{Average acc.} (\%) $\uparrow$} && \multicolumn{3}{c}{\emph{Forgetting rate.} (\%) $\downarrow$} \\
  \cmidrule{2-4} \cmidrule{6-8}
  & $N$=5 & $N$=10  & $N$=15 && $N$=5 & $N$=10  & $N$=15 \\
  \cmidrule{1-8}
  RGB only & 69.93 & 62.90 & 55.49 && 44.13 & 44.32 & 47.1\\
  \cmidrule{1-8}
  RGB-\textcolor{blue}{Edge} + \textcolor{red}{Early} & 70.94 & 63.90 & 55.28 && 42.4 & 42.13 & 45.80  \\
  RGB-\textcolor{blue}{Edge} + \textcolor{orange}{Intermediate} & \textbf{72.58} & \textbf{68.38} & \textbf{62.90} && \textbf{38.78} & \textbf{38.19} & \textbf{41.02}  \\
  \cmidrule{1-8}
     RGB-\textcolor{cyan}{Color} + \textcolor{red}{Early} & 73.58 & 64.57 & 53.56 && 37.825 & 42.86 & 46.15   \\
  RGB-\textcolor{cyan}{Color}+ \textcolor{orange}{Intermediate} & \textcolor{ForestGreen}{76.85} & \textcolor{ForestGreen}{69.94} & \textcolor{ForestGreen}{64.97} && \textcolor{ForestGreen}{33.15} & \textcolor{ForestGreen}{37.88} & \textcolor{ForestGreen}{39.79}  \\
  \cmidrule{1-8}
  RGB-\textcolor{blue}{Edge}-\textcolor{cyan}{Color} + \textcolor{red}{Early} & 69.99 & 63.33 & 56.34 && 42.35 & 44.17 & 46.24   \\
  RGB-\textcolor{blue}{Edge}-\textcolor{cyan}{Color}+ \textcolor{orange}{Intermediate} & \textbf{73.65} & \textbf{68.32} & \textbf{62.15} && \textbf{36.30} & \textbf{38.48} & \textbf{40.58} \\
  \bottomrule
\end{tabular}
}
  \label{tab:ablation_study}
\end{table*}
\textbf{Key Findings}. To the best of our knowledge, this work is the first to tackle the class incremental learning problem for the pill image domain, which is crucial and applicable for real-world pill recognition systems. Also, we empirically showed that the technique of intermediate fusion with the additional stream is superior to the early fusion technique. One plausible explanation for this effect is the flexibility of the fusion layer after it has been relocated to the intermediate stage. This allows the additional information to maintain its optimal performance for old tasks while learning to adapt to new tasks.\\
\\
\noindent \textbf{Limitations}. Though the proposed framework has superior performance over several state-of-the-art methods in CIL, it contains some limitations in different aspects. The new fusion layer at the intermediate phase might enlarge the model's size in terms of the number of parameters. Considering the scenario when a massive amount of tasks are encountered in the learning progress, the learning model could create sequences of abundant layers. This might create a side effect when too much memory is reserved for storing the model's parameters. Such reservation is unreasonable in a real-world deployment. Another restriction with the proposed framework is related to additional stream utilization. Apart from the traditional RGB stream, another information channel that is specific to the domain of usage might impose a disagreement with the original RGB channel. This requires a careful study of a different combination of streams accompanying the traditional stream to observe its effect.

\section{Conclusion}
This paper introduces the incremental learning capability to the traditional pill image classification systems. To this end, we propose a novel framework, namely Incremental Multi-stream Intermediate Fusion (IMIF) which integrates an additional stream of information to improve the performance of the single stream CIL method. We then devise CG-IMIF which utilizes IMIF along with a color histogram as guidance information. Our CG-IMIF is flexible and can be attached to any exemplar-based approach to improve the performance of the base ones. We experimentally show that CG-IMIF outperforms many existing state-of-the-art methods on the VAIPE-PCIL dataset. We hope our work would lay the foundation and could benefit several types of future research into the continual learning ability of intelligent machines in smart health applications.\\
\\
\textbf{Acknowledgements}. This work was funded by Vingroup Joint Stock Company (Vingroup JSC),Vingroup, and supported by Vingroup Innovation Foundation (VINIF) 
under
project code VINIF.2021.DA00128. Trong-Tung Nguyen was funded and and supported by the Master, PhD Scholarship Programme/Post-Doctoral Scholarship Programme of Vingroup Innovation Foundation (VINIF), Vingroup Big Data Institute (VinBigdata), code VINIF.2021.ThS.JVN.07.
\bibliographystyle{splncs04}
\bibliography{egbib}




% LNCS Homepage, \url{http://www.springer.com/lncs}. Last accessed 4
% \end{thebibliography}
\end{document}